\documentclass[acmtog, authorversio]{acmart}
\usepackage{multirow}
\usepackage{graphicx}
\graphicspath{{Figures/}}
\usepackage{subfigure}
\usepackage{flushend}
\usepackage{multicol} 
\usepackage{lipsum} 
\usepackage{float}
\usepackage{nopageno}

\AtBeginDocument{%
  \providecommand\BibTeX{{%
    \normalfont B\kern-0.5em{\scshape i\kern-0.25em b}\kern-0.8em\TeX}}}

\setcopyright{acmcopyright}
\copyrightyear{2018}
\acmYear{2018}
\acmDOI{XXXXXXX.XXXXXXX}

\acmConference[Conference acronym 'XX]{Make sure to enter the correct
  conference title from your rights confirmation emai}{June 03--05,
  2018}{Woodstock, NY}
\acmBooktitle{Woodstock '18: ACM Symposium on Neural Gaze Detection,
 June 03--05, 2018, Woodstock, NY} 
\acmPrice{15.00}
\acmISBN{978-1-4503-XXXX-X/18/06}

\usepackage{hyperref}
\begin{document}

\title{MRecGen: Multimodal Appropriate Reaction Generator}


\author{Jiaqi Xu}
\affiliation{%
  \institution{University of Leicester}
 \city{Leicester}
 \country{United Kingdom}
  }
\email{jiaqixu0626@gmail.com}

\author{Cheng Luo}
\affiliation{%
  \institution{Shenzhen University}
  \city{Shenzhen}
  \country{China}}
\email{luocheng2020@email.szu.edu.cn}

\author{Weicheng Xie}
\affiliation{%
  \institution{Shenzhen University}
  \city{Shenzhen}
  \country{China}}
\email{wcxie@szu.edu.cn}

\author{Linlin Shen}
\affiliation{%
  \institution{Shenzhen University}
  \city{Shenzhen}
  \country{China}}
\email{llshen@szu.edu.cn}

\author{Xiaofeng Liu}
\affiliation{%
  \institution{Hohai University}
  \city{Changzhou}
  \country{China}}
\email{xfliu@hhu.edu.cn}

\author{Lu Liu}
\affiliation{%
  \institution{University of Leicester}
 \city{Leicester}
 \country{United Kingdom}
  }
  \email{l.liu@leicester.ac.uk}

\author{Hatice Gunes}
\affiliation{%
  \institution{University of Cambridge}
  \city{Cambridge}
  \country{United Kingdom}}
\email{hatice.gunes@cl.cam.ac.uk}

\author{Siyang Song}
\authornote{Corresponding author.}
\affiliation{%
  \institution{University of Leicester \& University of Cambridge}
 \city{Leicester}
 \country{United Kingdom}
  }
  \email{ss1535@leicester.ac.uk}

\renewcommand{\shortauthors}{Xu et al.}

\begin{abstract}
Verbal and non-verbal human reaction generation is a challenging task, as different reactions could be appropriate for responding to the same behaviour. This paper proposes the first multiple and multimodal (verbal and nonverbal) appropriate human reaction generation framework that can generate appropriate and realistic human-style reactions (displayed in the form of synchronised text, audio and video streams) in response to an input user behaviour. This novel technique can be applied to various human-computer interaction scenarios by generating appropriate virtual agent/robot behaviours. Our demo is available at \url{https://github.com/SSYSteve/MRecGen}.
\end{abstract}






\maketitle


\section{Introduction}

\begin{figure*}
    \centering
    \includegraphics[width=1.9\columnwidth]{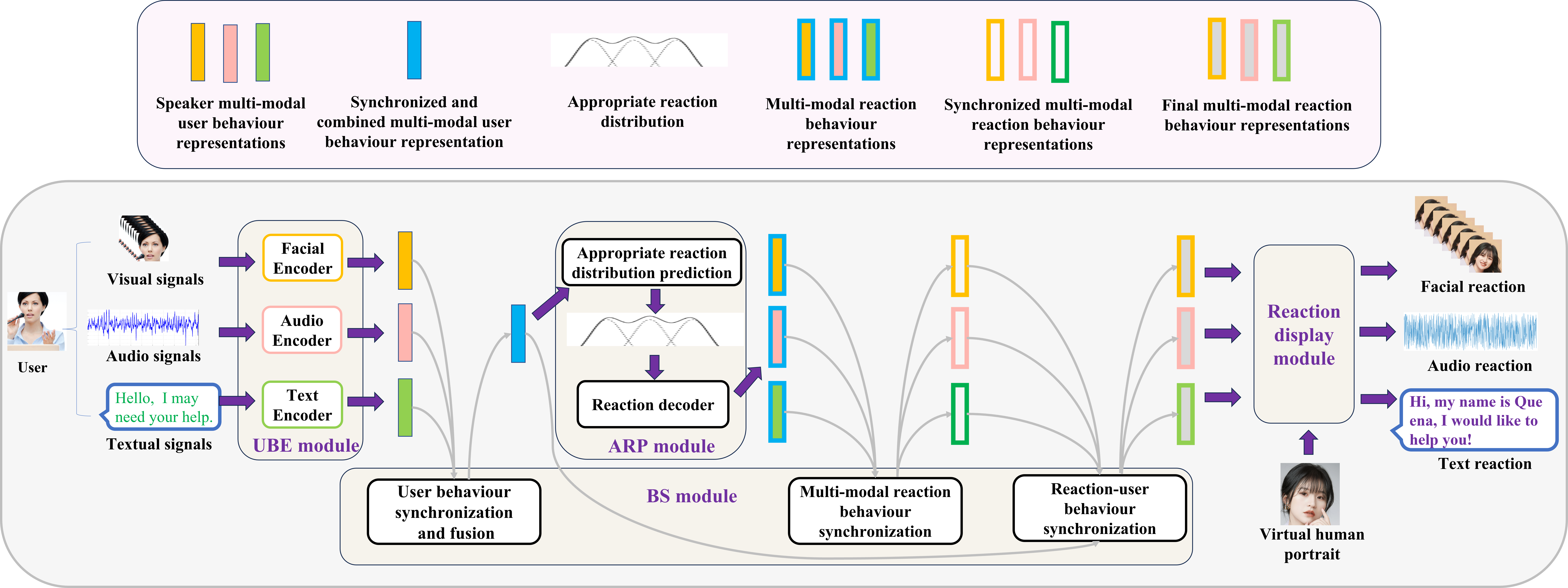}
    \caption{Overview of proposed MRecGen framework.}
    \label{fig:overview}
\end{figure*}


Automatic human behaviour reaction generation (ABRG) is a challenging task as multiple reactions (consisting of both verbal and non-verbal behaviours) could be \textit{appropriate} in response to the same behaviour expressed by conversational partner (defined as the speaker) \cite{song2023multiple}. This complexity and uncertainty arises from the interplay between individuals' internal cognitive processes, personal characteristics, and external environmental factors \cite{mehrabian1974approach,song2022learning,zhai2020sor,pandita2021psychological,song2022learning,shao2021personality}.



\begin{figure}
    \centering
    \includegraphics[width=1\columnwidth]{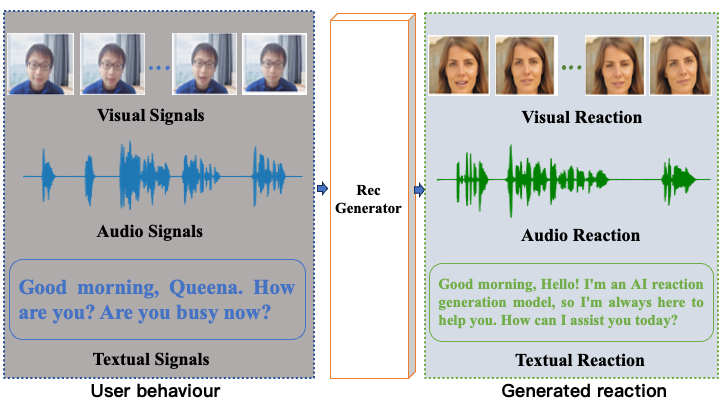} \caption{Example of the generated multi-modal reaction.}
    \label{fig:Reacation display}
\end{figure}

Recent advances in large language models (LLM) \cite{openai2023gpt4,chung2022scaling,chowdhery2022palm,biderman2023pythia,taylor2022galactica} result in powerful dialogue systems. Although these approaches, especially the GPT system \cite{openai2023gpt4}, can generate realistic verbal reactions (texts) in response to various textual inputs, they still lack the ability to generate non-verbal audio and visual reactions, which are integral components of authentic human behaviour. Although a few approaches have explored non-verbal reaction generation, such as facial reactions \cite{huang2017dyadgan,song2022learning,shao2021personality,ng2022learning} and gesture reactions \cite{barquero2022belfusion,jonell2020let,taras2020}, most of them rely on deterministic methods that aim to replicate the real reactions expressed by the corresponding listeners within a specific context. Consequently, these deterministic approaches suffer from 'one-to-many mapping' problem that arises from multiple different real reactions being triggered by the same speaker behavior. Thus, it is theoretically infeasible to develop a machine learning (ML) model capable of reproducing behavioural reactions from multiple subjects across diverse contexts. Consequently, a novel task called multiple appropriate reaction generation has been recently proposed \cite{song2023multiple,song2023react2023} and investigated \cite{xu2023reversible,luo2023reactface}.

In this paper, we propose and demonstrate the first fully automatic multiple appropriate human (verbal and non-verbal) behaviour reaction generation framework (called MRecGen). As illustrated in Fig. \ref{fig:overview}, the MRecGen consists of four main modules: a user behaviour encoding (UBE) module, an appropriate reaction prediction (ARP) module, a behaviour synchronisation (BS) module, and a reaction display (RD) module. Fig. \ref{fig:Reacation display} visualises that our framework can generate multiple appropriate, synchronised and realistic human verbal textual and non-verbal audio-facial behaviour reactions in response to a previous unseen speaker behaviour (i.e., can be either verbal behaviour only or verbal and non-verbal audio-facial behaviours) in dyadic interaction scenarios.



\section{MRecGen framework}
\label{sec:method}

The proposed MRecGen (demo) is an end-to-end deep learning framework consisting of four deep learning modules, which are introduced as follows:

\noindent \textbf{User behaviour encoding:} The UBE module is a multi-modal transformer which takes the raw multi-modal user behaviour (e.g., audio, text and visual behaviour) as the input, and then encodes them as a set of user behaviour latent representations.

\noindent \textbf{Appropriate reaction prediction:} The ARP module takes aligned and combined speaker behaviour representations (produced by the BS module) as the input, based on which it predicts a distribution representing multiple different but verbal and non-verbal reactions. These reactions are expected to be appropriate for responding to the input user behaviour. Finally, this module decodes multiple sets of appropriate reaction representations from the predicted distribution, where each set contains three latent representations describing the text, audio and facial behaviours of an appropriate reaction.

\noindent \textbf{Behaviour synchronisation:} This module conducts following operations: (i) synchronising multiple user behaviour representations generated by the UBE module, and then combine them as a single user behaviour representation; (ii) synchronising multiple reaction representations generated by ARP module, which represents the multi-modal reaction behaviour; and (iii) synchronising the synchronised multi-modal reaction representations with the synchronised and combined multi-modal user behaviour representation.

\noindent \textbf{Reaction display:} This module finally displays the generated reactions in the form of audio, text and face video.

\paragraph{Implementation details:} Our demo employs the architecture proposed in \cite{tsai2019multimodal} as the SBE module. The distribution learning strategy in ARP module is inherited from \cite{xu2023reversible}, where the speaker representation is transformed into graph representation to predict appropriate reaction distribution. The transformer-based multi-modal and inter-person behaviour synchronisation operations defined by the BS module are built based on the similar strategy proposed in \cite{luo2023reactface}. For the reaction display module, GPT4 \cite{openai2023gpt4}, Bark \footnote{\url{https://github.com/suno-ai/bark}}, and SadTalker \cite{zhang2023sadtalker} are employed to generate the final text, audio and facial video, respectively.

\section{Demo evaluation}
\label{sec:evaluation}

We recruited 58 volunteers (19 females and 39 males) online for the following two user studies, where each volunteer is asked to evaluate the performances of two tasks:
\begin{itemize}

    \item (i) Evaluating the demo's ability in generating reactions in response to users' audio-visual behaviours. Each volunteer is asked to watch five examples, where each example includes: (1) an audio-visual user clip; (2) a corresponding audio-visual-text human reaction clip; and (3) an audio-visual-text reaction clip generated by our demo.

    \item (ii) Evaluating the demo's ability in generating reactions in response to users' verbal textual behaviours. Each volunteer is asked to watch five examples, where each example includes: (1) an textual input by user; and (2) an audio-visual-text reaction clip generated by our demo.

\end{itemize}
We employ the widely-used Mean Opinion Scores (MOS) rating protocol \cite{streijl2016mean}, where users are required to give their ratings (1-5) on the following seven aspects for each video : (1) textual response appropriateness, (2) audio response appropriateness, (3) audio response smoothness, (4) facial reaction appropriateness, (5) facial reaction smoothness, (6) lip sync quality, and (7) video realism. The results show that volunteers feel that the reactions generated by our demo are good in all aspects (i.e.,  scores are above 3.0 for all 7 aspects). Particularly, our demo generated appropriate textual/facial reactions with high lip sync quality (their scores are above 3.4). The details of these ratings are provided in Supplementary Material.




\section{Conclusion}
\label{sec:conclusion}

This paper proposes the first multiple and multi-modal appropriate human behaviour reaction generation framework, and provides a well-trained model (demo) that can generate multiple appropriate, synchronised and realistic human textual, audio and facial behaviour reactions in response to user behaviours. 


\clearpage

\bibliographystyle{ACM-Reference-Format}
\bibliography{sample-base}


\begin{thebibliography}{22}


\ifx \showCODEN    \undefined \def \showCODEN     #1{\unskip}     \fi
\ifx \showDOI      \undefined \def \showDOI       #1{#1}\fi
\ifx \showISBNx    \undefined \def \showISBNx     #1{\unskip}     \fi
\ifx \showISBNxiii \undefined \def \showISBNxiii  #1{\unskip}     \fi
\ifx \showISSN     \undefined \def \showISSN      #1{\unskip}     \fi
\ifx \showLCCN     \undefined \def \showLCCN      #1{\unskip}     \fi
\ifx \shownote     \undefined \def \shownote      #1{#1}          \fi
\ifx \showarticletitle \undefined \def \showarticletitle #1{#1}   \fi
\ifx \showURL      \undefined \def \showURL       {\relax}        \fi
\providecommand\bibfield[2]{#2}
\providecommand\bibinfo[2]{#2}
\providecommand\natexlab[1]{#1}
\providecommand\showeprint[2][]{arXiv:#2}

\bibitem[Barquero et~al\mbox{.}(2022)]%
        {barquero2022belfusion}
\bibfield{author}{\bibinfo{person}{German Barquero}, \bibinfo{person}{Sergio
  Escalera}, {and} \bibinfo{person}{Cristina Palmero}.}
  \bibinfo{year}{2022}\natexlab{}.
\newblock \showarticletitle{BeLFusion: Latent Diffusion for Behavior-Driven
  Human Motion Prediction}.
\newblock \bibinfo{journal}{\emph{arXiv preprint arXiv:2211.14304}}
  (\bibinfo{year}{2022}).
\newblock


\bibitem[Biderman et~al\mbox{.}(2023)]%
        {biderman2023pythia}
\bibfield{author}{\bibinfo{person}{Stella Biderman}, \bibinfo{person}{Hailey
  Schoelkopf}, \bibinfo{person}{Quentin Anthony}, \bibinfo{person}{Herbie
  Bradley}, \bibinfo{person}{Kyle O'Brien}, \bibinfo{person}{Eric Hallahan},
  \bibinfo{person}{Mohammad~Aflah Khan}, \bibinfo{person}{Shivanshu Purohit},
  \bibinfo{person}{USVSN~Sai Prashanth}, \bibinfo{person}{Edward Raff},
  {et~al\mbox{.}}} \bibinfo{year}{2023}\natexlab{}.
\newblock \showarticletitle{Pythia: A suite for analyzing large language models
  across training and scaling}.
\newblock \bibinfo{journal}{\emph{arXiv preprint arXiv:2304.01373}}
  (\bibinfo{year}{2023}).
\newblock


\bibitem[Chowdhery et~al\mbox{.}(2022)]%
        {chowdhery2022palm}
\bibfield{author}{\bibinfo{person}{Aakanksha Chowdhery},
  \bibinfo{person}{Sharan Narang}, \bibinfo{person}{Jacob Devlin},
  \bibinfo{person}{Maarten Bosma}, \bibinfo{person}{Gaurav Mishra},
  \bibinfo{person}{Adam Roberts}, \bibinfo{person}{Paul Barham},
  \bibinfo{person}{Hyung~Won Chung}, \bibinfo{person}{Charles Sutton},
  \bibinfo{person}{Sebastian Gehrmann}, {et~al\mbox{.}}}
  \bibinfo{year}{2022}\natexlab{}.
\newblock \showarticletitle{Palm: Scaling language modeling with pathways}.
\newblock \bibinfo{journal}{\emph{arXiv preprint arXiv:2204.02311}}
  (\bibinfo{year}{2022}).
\newblock


\bibitem[Chung et~al\mbox{.}(2022)]%
        {chung2022scaling}
\bibfield{author}{\bibinfo{person}{Hyung~Won Chung}, \bibinfo{person}{Le Hou},
  \bibinfo{person}{Shayne Longpre}, \bibinfo{person}{Barret Zoph},
  \bibinfo{person}{Yi Tay}, \bibinfo{person}{William Fedus},
  \bibinfo{person}{Eric Li}, \bibinfo{person}{Xuezhi Wang},
  \bibinfo{person}{Mostafa Dehghani}, \bibinfo{person}{Siddhartha Brahma},
  {et~al\mbox{.}}} \bibinfo{year}{2022}\natexlab{}.
\newblock \showarticletitle{Scaling instruction-finetuned language models}.
\newblock \bibinfo{journal}{\emph{arXiv preprint arXiv:2210.11416}}
  (\bibinfo{year}{2022}).
\newblock


\bibitem[Huang and Khan(2017)]%
        {huang2017dyadgan}
\bibfield{author}{\bibinfo{person}{Yuchi Huang} {and} \bibinfo{person}{Saad~M
  Khan}.} \bibinfo{year}{2017}\natexlab{}.
\newblock \showarticletitle{Dyadgan: Generating facial expressions in dyadic
  interactions}. In \bibinfo{booktitle}{\emph{Proceedings of the IEEE
  Conference on Computer Vision and Pattern Recognition Workshops}}.
  \bibinfo{pages}{11--18}.
\newblock


\bibitem[Jonell et~al\mbox{.}(2020)]%
        {jonell2020let}
\bibfield{author}{\bibinfo{person}{Patrik Jonell}, \bibinfo{person}{Taras
  Kucherenko}, \bibinfo{person}{Gustav~Eje Henter}, {and}
  \bibinfo{person}{Jonas Beskow}.} \bibinfo{year}{2020}\natexlab{}.
\newblock \showarticletitle{Let's face it: Probabilistic multi-modal
  interlocutor-aware generation of facial gestures in dyadic settings}. In
  \bibinfo{booktitle}{\emph{Proceedings of the 20th ACM International
  Conference on Intelligent Virtual Agents}}. \bibinfo{pages}{1--8}.
\newblock


\bibitem[Kucherenko et~al\mbox{.}(2020)]%
        {taras2020}
\bibfield{author}{\bibinfo{person}{Taras Kucherenko}, \bibinfo{person}{Patrik
  Jonell}, \bibinfo{person}{Sanne van Waveren}, \bibinfo{person}{Gustav~Eje
  Henter}, \bibinfo{person}{Simon Alexandersson}, \bibinfo{person}{Iolanda
  Leite}, {and} \bibinfo{person}{Hedvig Kjellstr\"{o}m}.}
  \bibinfo{year}{2020}\natexlab{}.
\newblock \showarticletitle{Gesticulator: A Framework for Semantically-Aware
  Speech-Driven Gesture Generation}. In \bibinfo{booktitle}{\emph{Proceedings
  of the 2020 International Conference on Multimodal Interaction}} (Virtual
  Event, Netherlands) \emph{(\bibinfo{series}{ICMI '20})}.
  \bibinfo{publisher}{Association for Computing Machinery},
  \bibinfo{address}{New York, NY, USA}, \bibinfo{pages}{242–250}.
\newblock
\showISBNx{9781450375818}
\urldef\tempurl%
\url{https://doi.org/10.1145/3382507.3418815}
\showDOI{\tempurl}


\bibitem[Luo et~al\mbox{.}(2023)]%
        {luo2023reactface}
\bibfield{author}{\bibinfo{person}{Cheng Luo}, \bibinfo{person}{Siyang Song},
  \bibinfo{person}{Weicheng Xie}, \bibinfo{person}{Micol Spitale},
  \bibinfo{person}{Linlin Shen}, {and} \bibinfo{person}{Hatice Gunes}.}
  \bibinfo{year}{2023}\natexlab{}.
\newblock \showarticletitle{ReactFace: Multiple Appropriate Facial Reaction
  Generation in Dyadic Interactions}.
\newblock \bibinfo{journal}{\emph{arXiv preprint arXiv:2305.15748}}
  (\bibinfo{year}{2023}).
\newblock


\bibitem[Mehrabian and Russell(1974)]%
        {mehrabian1974approach}
\bibfield{author}{\bibinfo{person}{Albert Mehrabian} {and}
  \bibinfo{person}{James~A Russell}.} \bibinfo{year}{1974}\natexlab{}.
\newblock \bibinfo{booktitle}{\emph{An approach to environmental psychology.}}
\newblock \bibinfo{publisher}{the MIT Press}.
\newblock


\bibitem[Ng et~al\mbox{.}(2022)]%
        {ng2022learning}
\bibfield{author}{\bibinfo{person}{Evonne Ng}, \bibinfo{person}{Hanbyul Joo},
  \bibinfo{person}{Liwen Hu}, \bibinfo{person}{Hao Li}, \bibinfo{person}{Trevor
  Darrell}, \bibinfo{person}{Angjoo Kanazawa}, {and} \bibinfo{person}{Shiry
  Ginosar}.} \bibinfo{year}{2022}\natexlab{}.
\newblock \showarticletitle{Learning to Listen: Modeling Non-Deterministic
  Dyadic Facial Motion}. In \bibinfo{booktitle}{\emph{Proceedings of the
  IEEE/CVF Conference on Computer Vision and Pattern Recognition}}.
  \bibinfo{pages}{20395--20405}.
\newblock


\bibitem[OpenAI(2023)]%
        {openai2023gpt4}
\bibfield{author}{\bibinfo{person}{OpenAI}.} \bibinfo{year}{2023}\natexlab{}.
\newblock \bibinfo{title}{GPT-4 Technical Report}.
\newblock
\newblock
\showeprint[arxiv]{2303.08774}~[cs.CL]


\bibitem[Pandita et~al\mbox{.}(2021)]%
        {pandita2021psychological}
\bibfield{author}{\bibinfo{person}{Shailesh Pandita},
  \bibinfo{person}{Hari~Govind Mishra}, {and} \bibinfo{person}{Shagun Chib}.}
  \bibinfo{year}{2021}\natexlab{}.
\newblock \showarticletitle{Psychological impact of covid-19 crises on students
  through the lens of Stimulus-Organism-Response (SOR) model}.
\newblock \bibinfo{journal}{\emph{Children and Youth Services Review}}
  \bibinfo{volume}{120} (\bibinfo{year}{2021}), \bibinfo{pages}{105783}.
\newblock


\bibitem[Shao et~al\mbox{.}(2021)]%
        {shao2021personality}
\bibfield{author}{\bibinfo{person}{Zilong Shao}, \bibinfo{person}{Siyang Song},
  \bibinfo{person}{Shashank Jaiswal}, \bibinfo{person}{Linlin Shen},
  \bibinfo{person}{Michel Valstar}, {and} \bibinfo{person}{Hatice Gunes}.}
  \bibinfo{year}{2021}\natexlab{}.
\newblock \showarticletitle{Personality recognition by modelling
  person-specific cognitive processes using graph representation}. In
  \bibinfo{booktitle}{\emph{proceedings of the 29th ACM international
  conference on multimedia}}. \bibinfo{pages}{357--366}.
\newblock


\bibitem[Song et~al\mbox{.}(2022)]%
        {song2022learning}
\bibfield{author}{\bibinfo{person}{Siyang Song}, \bibinfo{person}{Zilong Shao},
  \bibinfo{person}{Shashank Jaiswal}, \bibinfo{person}{Linlin Shen},
  \bibinfo{person}{Michel Valstar}, {and} \bibinfo{person}{Hatice Gunes}.}
  \bibinfo{year}{2022}\natexlab{}.
\newblock \showarticletitle{Learning Person-specific Cognition from Facial
  Reactions for Automatic Personality Recognition}.
\newblock \bibinfo{journal}{\emph{IEEE Transactions on Affective Computing}}
  (\bibinfo{year}{2022}).
\newblock


\bibitem[Song et~al\mbox{.}(2023b)]%
        {song2023react2023}
\bibfield{author}{\bibinfo{person}{Siyang Song}, \bibinfo{person}{Micol
  Spitale}, \bibinfo{person}{Cheng Luo}, \bibinfo{person}{German Barquero},
  \bibinfo{person}{Cristina Palmero}, \bibinfo{person}{Sergio Escalera},
  \bibinfo{person}{Michel Valstar}, \bibinfo{person}{Tobias Baur},
  \bibinfo{person}{Fabien Ringeval}, \bibinfo{person}{Elisabeth Andre},
  {et~al\mbox{.}}} \bibinfo{year}{2023}\natexlab{b}.
\newblock \showarticletitle{REACT2023: the first Multi-modal Multiple
  Appropriate Facial Reaction Generation Challenge}.
\newblock \bibinfo{journal}{\emph{arXiv preprint arXiv:2306.06583}}
  (\bibinfo{year}{2023}).
\newblock


\bibitem[Song et~al\mbox{.}(2023a)]%
        {song2023multiple}
\bibfield{author}{\bibinfo{person}{Siyang Song}, \bibinfo{person}{Micol
  Spitale}, \bibinfo{person}{Yiming Luo}, \bibinfo{person}{Batuhan Bal}, {and}
  \bibinfo{person}{Hatice Gunes}.} \bibinfo{year}{2023}\natexlab{a}.
\newblock \showarticletitle{Multiple Appropriate Facial Reaction Generation in
  Dyadic Interaction Settings: What, Why and How?}
\newblock \bibinfo{journal}{\emph{arXiv e-prints}} (\bibinfo{year}{2023}),
  \bibinfo{pages}{arXiv--2302}.
\newblock


\bibitem[Streijl et~al\mbox{.}(2016)]%
        {streijl2016mean}
\bibfield{author}{\bibinfo{person}{Robert~C Streijl}, \bibinfo{person}{Stefan
  Winkler}, {and} \bibinfo{person}{David~S Hands}.}
  \bibinfo{year}{2016}\natexlab{}.
\newblock \showarticletitle{Mean opinion score (MOS) revisited: methods and
  applications, limitations and alternatives}.
\newblock \bibinfo{journal}{\emph{Multimedia Systems}} \bibinfo{volume}{22},
  \bibinfo{number}{2} (\bibinfo{year}{2016}), \bibinfo{pages}{213--227}.
\newblock


\bibitem[Taylor et~al\mbox{.}(2022)]%
        {taylor2022galactica}
\bibfield{author}{\bibinfo{person}{Ross Taylor}, \bibinfo{person}{Marcin
  Kardas}, \bibinfo{person}{Guillem Cucurull}, \bibinfo{person}{Thomas
  Scialom}, \bibinfo{person}{Anthony Hartshorn}, \bibinfo{person}{Elvis
  Saravia}, \bibinfo{person}{Andrew Poulton}, \bibinfo{person}{Viktor Kerkez},
  {and} \bibinfo{person}{Robert Stojnic}.} \bibinfo{year}{2022}\natexlab{}.
\newblock \showarticletitle{Galactica: A large language model for science}.
\newblock \bibinfo{journal}{\emph{arXiv preprint arXiv:2211.09085}}
  (\bibinfo{year}{2022}).
\newblock


\bibitem[Tsai et~al\mbox{.}(2019)]%
        {tsai2019multimodal}
\bibfield{author}{\bibinfo{person}{Yao-Hung~Hubert Tsai},
  \bibinfo{person}{Shaojie Bai}, \bibinfo{person}{Paul~Pu Liang},
  \bibinfo{person}{J~Zico Kolter}, \bibinfo{person}{Louis-Philippe Morency},
  {and} \bibinfo{person}{Ruslan Salakhutdinov}.}
  \bibinfo{year}{2019}\natexlab{}.
\newblock \showarticletitle{Multimodal transformer for unaligned multimodal
  language sequences}. In \bibinfo{booktitle}{\emph{Proceedings of the
  conference. Association for Computational Linguistics. Meeting}},
  Vol.~\bibinfo{volume}{2019}. NIH Public Access, \bibinfo{pages}{6558}.
\newblock


\bibitem[Xu et~al\mbox{.}(2023)]%
        {xu2023reversible}
\bibfield{author}{\bibinfo{person}{Tong Xu}, \bibinfo{person}{Micol Spitale},
  \bibinfo{person}{Hao Tang}, \bibinfo{person}{Lu Liu}, \bibinfo{person}{Hatice
  Gunes}, {and} \bibinfo{person}{Siyang Song}.}
  \bibinfo{year}{2023}\natexlab{}.
\newblock \showarticletitle{Reversible Graph Neural Network-based Reaction
  Distribution Learning for Multiple Appropriate Facial Reactions Generation}.
\newblock \bibinfo{journal}{\emph{arXiv preprint arXiv:2305.15270}}
  (\bibinfo{year}{2023}).
\newblock


\bibitem[Zhai et~al\mbox{.}(2020)]%
        {zhai2020sor}
\bibfield{author}{\bibinfo{person}{Xuesong Zhai}, \bibinfo{person}{Minjuan
  Wang}, {and} \bibinfo{person}{Usman Ghani}.} \bibinfo{year}{2020}\natexlab{}.
\newblock \showarticletitle{The SOR (stimulus-organism-response) paradigm in
  online learning: an empirical study of students’ knowledge hiding
  perceptions}.
\newblock \bibinfo{journal}{\emph{Interactive Learning Environments}}
  \bibinfo{volume}{28}, \bibinfo{number}{5} (\bibinfo{year}{2020}),
  \bibinfo{pages}{586--601}.
\newblock


\bibitem[Zhang et~al\mbox{.}(2023)]%
        {zhang2023sadtalker}
\bibfield{author}{\bibinfo{person}{Wenxuan Zhang}, \bibinfo{person}{Xiaodong
  Cun}, \bibinfo{person}{Xuan Wang}, \bibinfo{person}{Yong Zhang},
  \bibinfo{person}{Xi Shen}, \bibinfo{person}{Yu Guo}, \bibinfo{person}{Ying
  Shan}, {and} \bibinfo{person}{Fei Wang}.} \bibinfo{year}{2023}\natexlab{}.
\newblock \showarticletitle{SadTalker: Learning Realistic 3D Motion
  Coefficients for Stylized Audio-Driven Single Image Talking Face Animation}.
  In \bibinfo{booktitle}{\emph{Proceedings of the IEEE/CVF Conference on
  Computer Vision and Pattern Recognition}}. \bibinfo{pages}{8652--8661}.
\newblock


\end{thebibliography}

\newpage



\section*{Supplementary Material}

\begin{figure}[htbp]
    \centering
    \includegraphics[width=1\columnwidth]{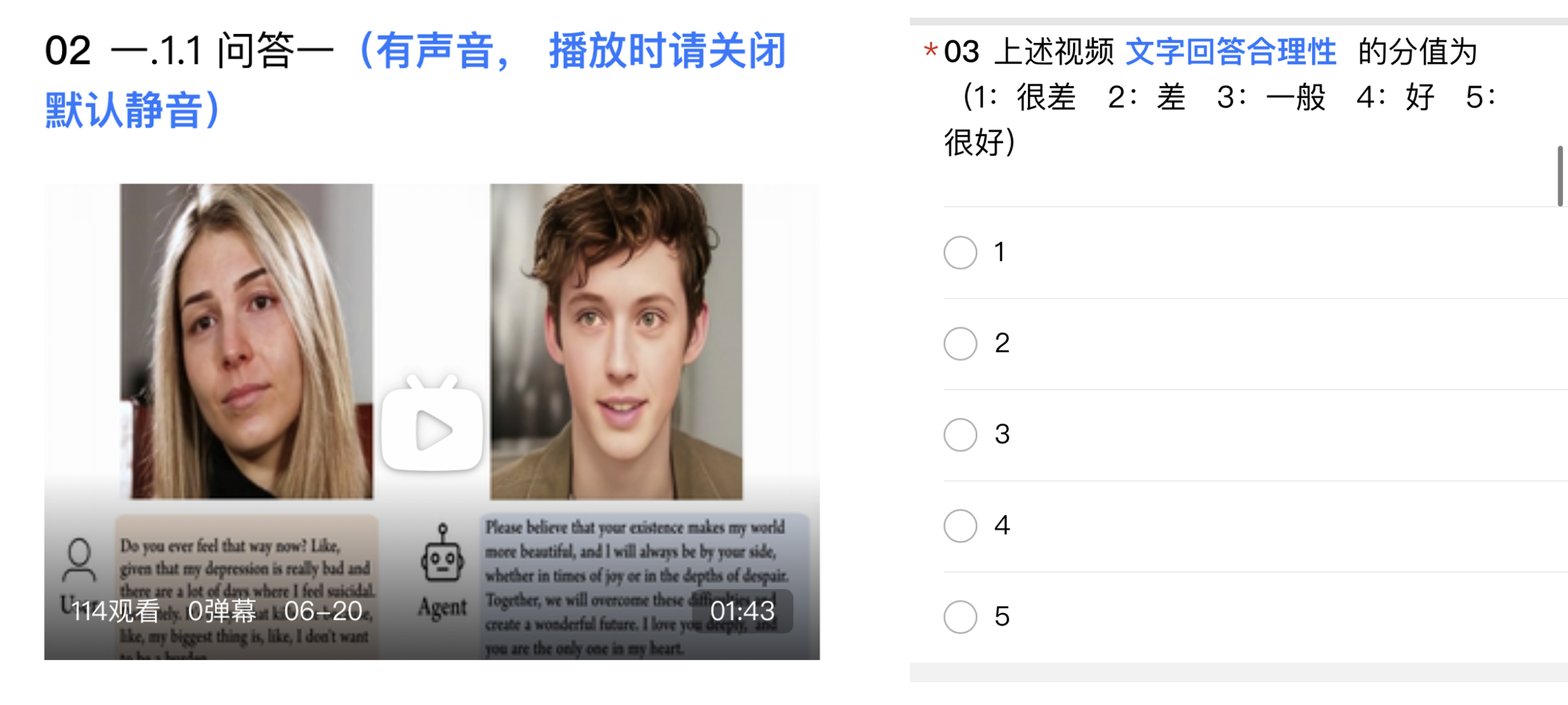} 
    \caption{Illustration of our user studies.}
    \label{fig:supple}
\end{figure}

\noindent We recruited 58 volunteers (including 19 females and 39 males, and all of them are Chinese) via the ‘Tencent Questionnaire’ platform for the two user studies. An example of the user study screenshot is displayed in Fig. \ref{fig:supple}. 

In the following, we provide the detailed user study results (including both gender dependent and independent results). Specifically, the gender independent results of the task 1 (i.e., each input is a human audio-visual clip) are reported in Table \ref{tab:user_study_1}, where the inter agreement (Cronbach's alpha) for rating GT and reactions generated by our MRenGen are 0.478 and 0.803, respectively. Meanwhile, the task 1 results achieved from female users and male users are reported in Table \ref{tab:user_study_1_female} and Table \ref{tab:user_study_1_male}, respectively.

The gender independent results of the task 2 (i.e., each input is the text) are reported in Table \ref{tab:user_study_2}, where the inter agreement (Cronbach's alpha) for rating reactions generated by our MRenGen is 0.825, respectively. Meanwhile, the task 2 results achieved from female users and male users are reported in Table \ref{tab:user_study_2_female} and Table \ref{tab:user_study_2_male}, respectively.

\begin{table}[htbp]
    \centering
    \begin{tabular}{l|cc}
    \toprule
      Aspect   & GT & MRecGen \\
    \midrule
        Textual response appropriateness &4.11 $\pm$ 0.77 &3.69 $\pm$ 0.82 \\
        Audio response appropriateness &4.18 $\pm$ 0.77  & 3.16 $\pm$  0.83\\
        Audio response naturalness &4.20 $\pm$ 0.80  &3.10 $\pm$ 0.98 \\
        Facial reaction appropriateness &4.24 $\pm$ 0.78 &3.44  $\pm$ 0.87\\
        Facial reaction naturalness &4.23  $\pm$  0.76 &3.31 $\pm$ 0.83 \\
        Lip sync quality &4.27 $\pm$ 0.76  &3.51 $\pm$ 0.94 \\
        Video realism &4.28 $\pm$ 0.74 &3.21  $\pm$  0.98\\
    \bottomrule
    \end{tabular}
    \caption{ User study results of the task 1 (average and standard deviation of Mean Opinion Scores), where the input is a human audio-visual clip. Higher is better, with the upper-bound of 5 and lower-bound of 0.}
    \label{tab:user_study_1}
\end{table}
\newpage

\begin{table}[htbp]
    \centering
    \begin{tabular}{l|cc}
    \toprule
      Aspect   & GT & MRecGen \\
    \midrule
        Textual response appropriateness &4.03 $\pm$ 0.78 &3.71 $\pm$ 0.82 \\
        Audio response appropriateness &4.13 $\pm$ 0.89  & 3.24 $\pm$  0.93\\
        Audio response naturalness &4.18 $\pm$ 0.76  &3.16 $\pm$ 0.90 \\
        Facial reaction appropriateness &4.21 $\pm$ 0.73 &3.16  $\pm$ 0.84\\
        Facial reaction naturalness &4.11  $\pm$  0.64 &3.11 $\pm$ 0.97 \\
        Lip sync quality &4.26 $\pm$ 0.71  &3.58 $\pm$ 0.78 \\
    \bottomrule
    \end{tabular}
    \caption{Gender dependent (female) user study results of the task 1 (average and standard deviation of Mean Opinion Scores), where the input is a human audio-visual clip. Higher is better, with the upper-bound of 5 and lower-bound of 0.}
    \label{tab:user_study_1_female}
\end{table}


\begin{table}[htbp]
    \centering
    \begin{tabular}{l|cc}
    \toprule
      Aspect   & GT & MRecGen \\
    \midrule
        Textual response appropriateness &4.15 $\pm$ 0.77 &3.68$\pm$ 0.82 \\
        Audio response appropriateness &4.29 $\pm$ 0.72  & 3.54 $\pm$  0.81\\
        Audio response naturalness &4.26 $\pm$ 0.76  &3.38 $\pm$ 0.77 \\
        Facial reaction appropriateness &4.17 $\pm$ 0.79 &3.17  $\pm$ 0.82\\
        Facial reaction naturalness &4.24  $\pm$  0.87 &3.10 $\pm$ 0.97 \\
        Lip sync quality &4.27 $\pm$ 0.78  &3.47 $\pm$ 1.00 \\
    \bottomrule
    \end{tabular}
    \caption{Gender dependent (male) user study results of the task 1 (average and standard deviation of Mean Opinion Scores), where the input is a human audio-visual clip. Higher is better, with the upper-bound of 5 and lower-bound of 0.}
    \label{tab:user_study_1_male}    
\end{table}

\begin{table}[htbp]
    \centering
    \begin{tabular}{l|c}
    \toprule
       Aspect   & MRecGen \\
    \midrule
        Textual response appropriateness  &3.80 $\pm$  0.89\\
        Audio response appropriateness  &3.37  $\pm$  0.90\\
        Audio response naturalness  &3.34 $\pm$  0.92 \\
        Facial reaction appropriateness  &3.64 $\pm$  0.79 \\
        Facial reaction naturalness  &3.63 $\pm$  0.87 \\
        Lip sync quality  &3.51 $\pm$ 0.87 \\
        Video realism  &3.37  $\pm$  0.82 \\
    \bottomrule
    \end{tabular}
    \caption{User study results of the task 2 (average and standard deviation of Mean Opinion Scores), where the input is the text. Higher is better, with the upper-bound of 5 and lower-bound of 0.}
    \label{tab:user_study_2}
\end{table}
\newpage

\begin{table}[htbp]
    \centering
    \begin{tabular}{l|c}
    \toprule
       Aspect   & MRecGen \\
    \midrule
        Textual response appropriateness  &3.82 $\pm$  0.85\\
        Audio response appropriateness  &3.58  $\pm$  0.78\\
        Audio response naturalness  &3.55 $\pm$  0.91 \\
        Facial reaction appropriateness  &3.47 $\pm$  0.79 \\
        Facial reaction naturalness  &3.42 $\pm$  0.85 \\
        Lip sync quality  &3.58 $\pm$ 0.85 \\
        Video realism  &3.39  $\pm$  0.71\\


    \bottomrule
    \end{tabular}
    \caption{ Female User study results of the task 2 (average and standard deviation of Mean Opinion Scores), where the input is the text. Higher is better, with the upper-bound of 5 and lower-bound of 0.}
    \label{tab:user_study_2_female}
\end{table}

\begin{table}[htbp]
    \centering
    \begin{tabular}{l|c}
    \toprule
       Aspect   & MRecGen \\
    \midrule
        Textual response appropriateness  &3.79 $\pm$  0.90\\
        Audio response appropriateness  &3.67  $\pm$  0.80\\
        Audio response naturalness  &3.67 $\pm$  0.84 \\
        Facial reaction appropriateness  &3.32 $\pm$  0.94 \\
        Facial reaction naturalness  &3.31 $\pm$  0.94 \\
        Lip sync quality  &3.46 $\pm$ 0.87 \\
        Video realism  &3.36  $\pm$  0.88 \\


    \bottomrule
    \end{tabular}
    \caption{ Male User study results of the task 2 (average and standard deviation of Mean Opinion Scores), where the input is the text. Higher is better, with the upper-bound of 5 and lower-bound of 0.}
    \label{tab:user_study_2_male}
\end{table}

\end{document}